\documentclass[letterpaper,11pt]{article}

\usepackage{amssymb}
\usepackage{amsmath}
\usepackage{amsthm}
\usepackage{graphicx}
\usepackage{authblk}
\usepackage{xcolor}

\usepackage{float}

\usepackage{booktabs,siunitx}
\usepackage{multirow}

\usepackage{fullpage}
\usepackage{bm}

\usepackage[utf8x]{inputenc}

\usepackage{color}
\usepackage{algpseudocode, algorithm, algorithmicx}

\newcommand\amsclass[1]%
  {\hspace{2mm}
   \textbf{AMS subject classifications. }#1
}

\usepackage{romannum}

\title{A Neural Network Approach for Homogenization of Multiscale Problems}
\author{Jihun Han\thanks{jihun.han@dartmouth.edu} and Yoonsang Lee\thanks{yoonsang.lee@dartmouth.edu}}
\affil{Department of Mathematics, Dartmouth College}
\date{}

\begin{document}
\pagenumbering{arabic}

\maketitle
\begin{abstract}
We propose a neural network-based approach to the homogenization of multiscale problems. The proposed method uses a derivative-free formulation of a training loss, which incorporates Brownian walkers to find the macroscopic description of a multiscale PDE solution. Compared with other network-based approaches for multiscale problems, the proposed method is free from the design of hand-crafted neural network architecture and the cell problem to calculate the homogenization coefficient.
The exploration neighborhood of the Brownian walkers affects the overall learning trajectory. We determine the bounds of micro- and macro-time steps that capture the local heterogeneous and global homogeneous solution behaviors, respectively, through a neural network. The bounds imply that the computational cost of the proposed method is independent of the microscale periodic structure for the standard periodic problems. We validate the efficiency and robustness of the proposed method through a suite of linear and nonlinear multiscale problems with periodic and random field coefficients.
\end{abstract}

%\begin{keyword}
%
%\end{keyword}

\amsclass{65N99, 65C05, 68T07}

\section{Introduction}\label{sec:Introduction}
A wide range of problems in science and engineering involve multiscale nature. The underground of the earth is often described as heterogeneous material with spatial scales ranging from a few millimeters to kilometers \cite{rock}. The fluid motion of the oceans involves spatiotemporal scales from a few seconds to seasonal time scales and from small eddies of centimeters to zonal jets \cite{GFD}. In particular, the dynamics of sea ice thickness and momentum is characterized by microscale heterogeneous properties of saline ice, while its global properties have long-lasted (few months) and long-ranged (thousands of kilometers) correlations \cite{msseaice}. 

In the numerical simulation of multiscale problems, standard methods suffer from challenges in resolving all active scales due to their tremendously high and long computational cost and time. For computational efficiency and reliable prediction, many research efforts have focused on the averaged or macroscopic description of the multiscale solutions. Examples include large eddy simulation (LES) in turbulence modeling \cite{LES} and averaging methods of dynamical systems \cite{averaging}, to name a few.  Despite successful applications of the aforementioned methods in many applications, it still remains  a challenge to close the macroscopic equation in general as there are intertwined coupling through the diverse range of scales. 

For a certain class of problems, there are several numerical methods to approximate the  large-scale macroscopic behaviors of the multiscale problem. 
The asymptotic-preserving (AP) method \cite{AP} is a class of numerical methods that resolve the asymptotic perturbations of multiscale problems. Another class of methods, such as the heterogeneous multiscale method \cite{HMM} based on the homogenization theory of periodic structures or ergodic systems \cite{BLP}, provides a rigorous large-scale model approximated through a suite of local cell problems. To overcome the assumption of strong scale separation, a hierarchical approach has also been proposed for advection-diffusion problems \cite{lee2016Advdiff}. Instead of the averaged or homogenized solution, the generalized multiscale finite element method \cite{GMsFEM} aims for multiscale problems without scale separation over complex domains, which achieves a computational gain through offline calculations of multiscale basis functions.

%\paragraph{Neural network-based approaches for solving PDEs}
For the past few years, neural network-based methods have been proposed for solving partial differential equations (PDEs) \cite{raissi2019PINN,han2018solving, sirignano2018dgm}.
These methods can avoid the complex mesh generation process, which can be challenging for high-dimensional problems. The solution of a PDE can be found through the training process of the neural network by minimizing the loss of objective function that measures how well a neural network satisfies the PDE. Physics-informed neural networks (PINNs) 
\cite{raissi2019PINN} or deep Galerkin method (DGM) \cite{sirignano2018dgm} consider the direct PDE residual at each point within the domain. Deep Ritz method (DRM) \cite{yu2018deep} reformulates an elliptic PDE as an equivalent energy minimization problem to train a neural network. The approach proposed in \cite{han2018solving} estimates the solution of parabolic PDEs at a single point using a neural network tailored to the time discretization of the equivalent backward stochastic differential equation.

%\paragraph{Limitations and challenges in solving multiscale PDEs using NN}
Despite the remarkable accomplishments in solving a wide range of PDEs, neural network-based methods often face hurdles in solving multiscale problems with slow convergence rates or inaccurate approximation. Recent works have attempted to comprehend such unfavorable training trajectories. One direction addresses the intrinsic behavior of training neural networks, which is ineffective for learning functions involving a diverse frequency spectrum. The works in \cite{xu2019frequency,rahaman2019spectral,xu2019training} show that the gradient-based training process of standard multilayer perceptron (MLP) has spectral bias as the neural network tends to learn low frequencies first while it requires a longer time to fit high frequencies (called F-principle in \cite{xu2019frequency}). 

To overcome such limitations of standard MLPs for solving multiscale problems, many research efforts have focused on designing new architecture or modified ingredients of neural networks. The work in \cite{cai2019multi} proposed a neural network architecture with an input scaling treatment and a specialized activation function for converting the high-frequency components to low-frequency ones preferable to learning. The work in \cite{jagtap2020adaptive} introduced adaptive activation functions with trainable scaling factors, which improves the learning capability as it changes the topology of the loss function in an optimization process dynamically. Another work in \cite{wang2021eigenvector} demonstrated that multiple Fourier feature embedding of inputs allows a standard MLP to learn diverse frequency components efficiently, and the authors in \cite{han2021hierarchical} proposed that hierarchical organization of learning frequencies using the Fourier feature embedded networks 
improves the training performance. 

Another direction to use neural networks for multiscale problems examines the impediments to optimizing the multiple objective loss functions comprising the governing differential equation and initial/boundary conditions. The work in \cite{wang2021understanding} demonstrates that the imbalance of different loss components in the magnitude of the gradients could degrade the overall training process. The other work in \cite{wang2022and} estimates the convergence rate of different loss components in terms of the neural tangent kernel (NTK) \cite{jacot2018neural} and addresses the discrepancy in convergence rates as a fundamental reason for degradation in training performance. In these works, adaptive weighting on loss components was proposed to mitigate each corresponding imbalance. We note that the two main directions we introduce are not mutually exclusive, and combining two approaches could improve training performance in solving multiscale problems even more \cite{wang2021eigenvector}.

%\paragraph{Derivative-free loss approach} 
%\yl{Majda worked on walkers for homogenization problem; a PhD student at Princeton back in 199x; need to find the reference}.
This work proposes a neural network approach to approximate the homogenized solutions of multiscale elliptic problems. Our work is different from other neural network-based approaches for multiscale problems that rely on the computation of the homogenized coefficient. Instead, our proposed method is non-intrusive capturing the homogenized solution without knowing the homogenized coefficient or calculating it through cell problems. The proposed method utilizes the derivative-free loss method (DFLM) for solving PDEs \cite{han2020derivative}. The idea of the DFLM to avoid the expensive derivative calculations is a stochastic formulation of the PDE solution in the spirit of Feynman-Kac formula. In particular, the solution is represented as the expectation of a martingale process driven by standard Brownian motion. That is, the solution at a point $\bm{x}$ is an average of solution-related information on its Brownian motion neighborhood $\{\bm{B}_{t}:0\leq t \leq \Delta t, \bm{B}_0=\bm{x}\}$ for a given time duration.
This characteristic is distinct from other methods that use pointwise residual loss function in which values at nearby points are learned passively through a neural network.
We utilize such neighborhood-observing characteristic of the DFLM to extract readily learnable macroscopic information from complex heterogeneous properties.

The DFLM adopts a form of reinforcement learning where the target value of the neural network at each training point is iteratively updated, reflecting the information on its neighborhood. We address the size of the neighborhood as an important ingredient for learning homogenized features as it determines how much the heterogeneous properties modeled in the given equation are fed into the neural network in each training iteration. 
From the trade-off between the amount of information and computational efficiency, we propose an effective size of the neighborhood in terms of micro- and macro-time steps of Brownian motion. 
In particular, our method demands constant computational cost in standard periodic coefficient problems regardless of the given scale. 
Apart from the previous methods, our method does not require a well-designed neural network architecture, but a standard multilayer perceptron (MLP) is sufficiently trained to estimate the homogenized solutions. 
Also, we do not compute the homogenized coefficient referred to the homogenization theory, but the standard DFLM with proposed time steps efficiently approximates the homogenized solutions.

The rest of the paper is organized as follows. Section \ref{sec:DFLM} reviews the DFLM and discusses its characteristics with the comparison of other neural network-based methods. In section \ref{sec:proposed_method}, we propose the effective micro- and macro-time stepping in the DFLM for learning homogenized solutions of multiscale problems with supporting numerical validations. Section \ref{sec:Numerical Experiments} provides numerical experiments of linear and nonlinear multiscale problems with periodic and random field coefficients verifying the efficacy of the proposed method. Finally, we conclude with discussions about the limitation and future directions of the current study in Section \ref{sec:discussion}.

% SECTION2
\section{Neural network-based methods for solving PDEs}\label{sec:DFLM} 
In this section, we briefly review the key ideas of neural network-based methods for solving PDEs, including the Physics-informed neural network and the derivative-free loss method (DFLM). The latter method becomes the backbone of the proposed method in finding the homogenized solution of the multiscale problems using neural networks. The DFLM and its extension to multiscale problems can be applied to elliptic and parabolic equations. For simplicity of delivering the key idea of the proposed method, we focus on boundary value problems in the current study. 

Let us have a boundary value problem in a bounded domain $\Omega \in \mathbb{R}^{d}$,
\begin{align}\label{eq:PDE_general}
\begin{split}
\mathcal{N}[u](\bm{x}) &= f(\bm{x}), ~~ \bm{x} \in \Omega, \\
\mathcal{B}[u](\bm{x}) &= g(\bm{x}), ~~ \bm{x} \in \partial \Omega,
\end{split}
\end{align}
where $\mathcal{N}$ and $\mathcal{B}$ are the differential operator and boundary condition operator, respectively. 
A neural network representation of the solution $u$ is efficient for obtaining derivatives of neural networks, which allows one to represent the law of physics written in the differential equation without the need for operator discretization. 
For a network representation of the solution parameterized by $\bm{\theta}$, $u(\bm{x};\bm{\theta})$, the approximation is obtained by searching the optimal parameters $\bm{\theta}$ that minimizes the loss function $\mathcal{L}(\bm{\theta})$. Each network-based method can be categorized by the way to define the loss function. Once the loss function is defined, it is typically solved by a gradient descent method
\begin{equation}
\bm{\theta}_n = \bm{\theta}_{n-1} - \alpha \nabla_{\bm{\theta}} \mathcal{L}(\bm{\theta}_{n-1}), ~~\alpha > 0,
\end{equation}
where $\alpha$ is a learning parameter.

Another common characteristic of neural network-based methods is the randomized approach to evaluating integrals related to the loss function. Neural network-based methods treat the domain as a set of data points and define the loss function to minimize the disparity between a neural network and the PDE solution on the dataset. In the context of the stochastic gradient descent method (SGD), the data points for evaluating the loss function can be resampled at each iteration. This approach enables cost-efficient computation of the loss function and its gradient at each iteration, which becomes beneficial for high-dimensional problems.

\subsection{Derivative-based loss functions}
Different loss functions have been proposed either directly from a given PDE or using an equivalent formulation of the PDE.
Physics-informed neural network (PINN) \cite{raissi2019PINN} or deep Galerkin method (DGM) \cite{sirignano2018dgm} design the loss function $\mathcal{L}(\bm{\theta})$ to measure the direct PDE residual in both interior sample points $\left\{\bm{x}_r^{i}\right\}_{i=1}^{N_r}$ and the boundary sample points $\left\{\bm{x}_b^{i}\right\}_{i=1}^{N_b}$,
\begin{equation}
\mathcal{L}(\bm{\theta}) = \frac{\lambda_{\Omega}}{N_r}\sum \limits_{i=1}^{N_r}\left|\mathcal{N}[u(\cdot;\bm{\theta})](\bm{x}_r^{i})-f(\bm{x}_r^{i}) \right|^2 + \frac{\lambda_{\partial\Omega}}{N_b} \sum\limits_{i=1}^{N_b}\left|\mathcal{B}[u(\cdot;\bm{\theta})](\bm{x}_b^{i})-g(\bm{x}_b^{i}) \right|^2.
\end{equation}
Here, the derivatives of a neural network involved in the differential operator $\mathcal{N}$ and possibly in the boundary operator $\mathcal{B}$ are computed through the automatic differentiation (AD) or efficient Monte-Carlo approximations.
The parameters $\lambda_{\Omega}$ and $\lambda_{\partial \Omega}$ are chosen as hyperparameters or vary adaptively during the training process \cite{wang2021understanding, wang2022and}. For elliptic problems, deep Ritz method (DRM) \cite{yu2018deep} transforms an elliptic PDE to an equivalent variational formulation. For instance, the Poisson problem with Dirichlet boundary condition (i.e., $\mathcal{N}[u]=-\Delta u$, $\mathcal{B}[u]=g(\bm{x})$) is formulated as the following variational problem
\begin{equation}\label{eq:variational_form}
\min \limits_{u \in \mathcal{U}}I(u):= \int_{\Omega}\left(\frac{1}{2}|\nabla u(\bm{x})|^{2}-f(\bm{x})u(\bm{x})  \right)d\bm{x},~~ \mathcal{U} = \{u : u(\bm{x})=g(\bm{x}) ~~\textrm{on}~~ \partial \Omega \},
\end{equation}
and the neural network $u(\bm{x};\bm{\theta})$ is trained to solve the variational form $I(u)$ using the loss function
\begin{equation}
\mathcal{L}(\bm{\theta}) = \frac{1}{N_r}\sum \limits_{i=1}^{N_r}\left(\frac{1}{2}|\nabla u(\bm{x}_r^{i};\bm{\theta})|^{2}-f(\bm{x}_r^{i})u(\bm{x}_r^{i};\bm{\theta})  \right) + \frac{\beta}{N_b} \sum\limits_{i=1}^{N_b}\left|u(\bm{x}_b^{i};\bm{\theta})-g(\bm{x}_b^{i}) \right|^2.
\end{equation}
The second term on the right-hand side is understood as the penalty term to enforce the neural network in the functional space $\mathcal{U}$ of the boundary condition.

\subsection{Derivative-free loss method}
The derivative-free loss method (DFLM) proposed in \cite{han2020derivative} avoids the derivative of the function with the spatial domain variables. It reformulates a PDE to a martingale representation that describes the interrelation between a point and its neighborhood to represent a solution at the point. This formulation is different from the methods that use point-wise evaluation of the residual, such as PINN, in which learning occurs solely at each point, and the communication among nearby points is achieved passively through the network. 
Moreover, the DFLM adopts bootstrapping in the context of reinforcement learning that alternatively and gradually improves a neural network and corresponding target values toward the PDE solution. We note that this approach is different from the supervised learning methods that seek optimal neural network parameters within a fixed topology of a loss function.

To explain the main idea of the derivative-free loss formulation of the DFLM, we consider a class of elliptic PDEs of unknown function $u(\bm{x}) \in \mathbb{R}$ of the following form: 
\begin{equation}\label{eq:Quasi-linear elliptic}
\mathcal{N}[u](\bm{x}):= \frac{1}{2}\Delta u(\bm{x}) + \bm{V}\cdot \nabla u(\bm{x}) - G = 0, ~\textrm{in}~ \Omega \subset \mathbb{R}^d.
\end{equation}
Here $\bm{V}=\bm{V}(\bm{x},u(\bm{x})) \in \mathbb{R}^d$ is the advection velocity and $G=G(\bm{x}, u(\bm{x})) \in \mathbb{R}$ is the force term, which can depend on the unknown function $u$. 
The DFLM utilizes the theoretical connection between the stochastic process and a PDE to guide a neural network to learn the PDE solution, which is demonstrated as the following equivalence;
\begin{itemize}
\item $u:\Omega \rightarrow \mathbb{R}$ is a solution of Eq.~\eqref{eq:Quasi-linear elliptic}. \\
\vspace{-0.5cm}\item the stochastic process $q(t;u, \bm{x}, \{\bm{X}_s\}_{0\leq s\leq t}) \in \mathbb{R}$ defined as 
\begin{equation}\label{eq:stochastic_process}
\begin{split}
q(t;u, \bm{x},\{\bm{X}_s\}_{0\leq s\leq t}) := u(\bm{X}_t) - \int_0^t G(\bm{X}_s,u(\bm{X}_s))ds, \hspace{50mm}\\
\text{where}~ \bm{X}_t \in \mathbb{R}^d  \text{ is a soltuion of the SDE }~ 
d\bm{X}_t = \bm{V}(\bm{X}_t, u(\bm{X}_t))dt + d\bm{B}_t, \bm{X}_0=\bm{x}, \\
(\bm{B}_t:\text{standard Brownian motion in } \mathbb{R}^d )
\end{split}
\end{equation}
satisfy the martingale property
\begin{align}\label{eq:martingale}
\begin{split}
u(\bm{x})&=q(0;u,\bm{x}, \bm{X}_0)=\mathbb{E}\left[q(t;u,\bm{x}, \{\bm{X}_s\}_{0\leq s\leq t}) | \bm{X}_0=\bm{x}\right] \\
&= \mathbb{E}\left[u(\bm{X}_t) - \int_{0}^{t}G(\bm{X}_s, u(\bm{X}_s)) ds \middle |\bm{X}_0=\bm{x}  \right], ~~\textrm{for all}~ \bm{x}\in \Omega,~\textrm{and}~ t>0.
\end{split}
\end{align}
\end{itemize}
We note that the martingale property, Eq.~\eqref{eq:martingale}, holds for arbitrary time $t>0$ and any stopping time $\tau$ as well by the optional stopping theorem \cite{karatzas2012brownian}. In particular, the choice of the stopping time $\tau=\inf \{s: \bm{X}_s \notin \Omega \}$ induces the well-known Feynman-Kac formula for the PDE. Moreover, the Eq.~\eqref{eq:martingale} is also known as the Bellman equation in the context of Markov reward process, which appears in the value function estimation in reinforcement learning.

The classical Monte-Carlo methods based on the Feynman-Kac formula estimate the PDE solution at a single point independently from realizations of the stochastic process $\bm{X}_t$ until it exits from the given domain. On the other hand, the DFLM employs a neural network $u(\bm{x};\bm{\theta})$ to approximate the PDE solution over the domain at once, which is guided to satisfy the martingale property Eq.~\eqref{eq:martingale} within a short period of time, say $\Delta t$, rather than the time duration of complete trajectories until $\bm{X}_t$ is out of the domain. The corresponding loss functional for training a neural network is 
\begin{equation}\label{eq:loss_interior_continuous}
\mathcal{L}^{\Omega}(\bm{\theta}) = \frac{1}{N_r}\sum \limits_{i=1}^{N_r}\left|u(\bm{x}_i;\bm{\theta})-\widetilde{\mathbb{E}}\left[u(\bm{X}_{\Delta t};\bm{\theta}) - \int_{0}^{\Delta t}G(\bm{X}_s, u(\bm{X}_s;\bm{\theta})) ds \middle |\bm{X}_0=\bm{x}_i  \right] \right|^2,
\end{equation}
where $\{\bm{x}_i\}_{i=1}^{N_r}$ are sampling points in the interior of the domain and $\widetilde{\mathbb{E}}$ reads as an empirical mean using $N_s$ samples. We note that the boundary condition can be imposed on the stochastic process $\bm{X}_t$. For the Dirichlet boundary condition, $u(\bm{x})=g(\bm{x})$ on $\partial \Omega$, $\bm{X}_t$ exiting the domain during the small time $\Delta t$ is considered to cling onto the exit position on the boundary $\partial \Omega$ and the value of the neural network is replaced by the given boundary value at the position. In this manner, the information of the exact solution flows from the boundary into the interior of the domain. To enhance such information available on the boundary, the additional loss term 
\begin{equation}\label{eq:loss_boundary_dirichlet}
\mathcal{L}^{\partial \Omega}(\bm{\theta}) = \frac{1}{N_b}\sum\limits_{j=1}^{N_b}|u(\bm{x}_j;\bm{\theta})-g(\bm{x}_j) |^2, ~~\{\bm{x_j}\}_{j=1}^{N_b} \subset \partial \Omega,
\end{equation}
could be included in the loss functional. Moreover, the homogeneous Neumann condition can be imposed by reflecting $\bm{X}_t$ in the normal direction on the boundary $\partial \Omega$ modeling no-flux constraint.
A gradient descent method is applied to minimize the loss $\mathcal{L}^{\Omega}(\bm{\theta})$ and, in particular, bootstrapping approach is used as the target of the neural network at $\bm{x}_i$ (i.e., expectation component) is pre-evaluated using the current state of parameters $\bm{\theta}$. The $n$-th iteration step for updating the parameters $\bm{\theta}_n$ is summarized as 
\begin{equation}
\bm{\theta}_n = \bm{\theta}_{n-1} - \alpha \nabla\mathcal{L}_n(\bm{\theta}_{n-1}),
\end{equation}
\begin{equation}
\mathcal{L}_n(\bm{\theta}) :=  \frac{1}{N_r}\sum \limits_{i=1}^{N_r}\left|u(\bm{x}_i;\bm{\theta})-\widetilde{\mathbb{E}}\left[u(\bm{X}_{\Delta t};\bm{\theta}_{n-1}) - \int_{0}^{\Delta t}G(\bm{X}_s, u(\bm{X}_s;\bm{\theta}_{n-1})) ds \middle |\bm{X}_0=\bm{x}_i  \right] \right|^2.
\end{equation}
The learning rate $\alpha$ could be tuned at each step and the gradient step can also be optimized by taking the previous step into account such as Adam optimization \cite{kingma2014adam}. Moreover, the sampling point $\{\bm{x}_i\}_{i=1}^{N_r}$ could be chosen randomly at every iteration with a stochastic gradient method, and the standard sampling is the uniform distribution on $\Omega$. The original DFLM work \cite{han2020derivative} proposed a different sampling approach reflecting the physical conditions imposed on the differential equation; $N_r$ numbers of stochastic walkers $\bm{X}^{(i)}_t$, $i=1,2, \cdots N_r$, are moving ($\bm{X}_t^{(i)}\sim \bm{X}_t$) around the domain and the walkers' location at the discrete time step $t_n$, $\{\bm{X}^{(i)}_{t_n}\}_{i=1}^{N_r}$, are used as the training sample at $n$-th iteration. The sampling method can be understood as an importance sampling for Monte-Carlo integration of continuous loss functional, where its performance depends on applications. To moderate the nontrivial behavior of $\bm{X}_t$, an alternative martingale process $\tilde{q}(t;u,\bm{x},\{\bm{B}_s\}_{0\leq s\leq t})$ with standard Brownian motion $\bm{B}_t$ is proposed as 

\begin{equation}\label{eq:Brownian_martingale}
\begin{split}
\tilde{q}(t;u, \bm{x},\{\bm{B}_s\}_{0\leq s\leq t}) := 
 \left(u(\bm{B}_t) - \int_0^t G(\bm{B}_s,u(\bm{B}_s))ds\right) \mathcal{D}(\bm{V},u,t), \hspace{28mm}\\
 \text{where}~~\mathcal{D}(\bm{V},u,t)=\exp \left(\int^t_0\bm{V}(\bm{B}_s, u(\bm{B}_s))\cdot d\bm{B}_s-\frac{1}{2}\int^{t}_{0}|\bm{V}(\bm{B}_s, u(\bm{B}_s))|^2 ds \right).
\end{split}
\end{equation}
Here, the additional exponential factor $\mathcal{D}(\bm{V}, u,t)$ compensates the removal of the drift effect in $\bm{X}_t$. We call $\mathcal{R}(G,u,t):=\int_0^t G(\bm{B}_s,u(\bm{B}_s))ds$ and $\mathcal{D}(\bm{V},u,t)$ as \textit{reward} and \textit{discount}, respectively, regarding the context of the Bellman equation for the Markov reward process. The use of the alternative martingale allows the standard Brownian walkers to explore the domain regardless of the form of the given PDE. The loss functional corresponding to the $\tilde{q}$-martingale is 
\begin{equation}\label{eq:loss_interior_continuous_ver2}
\mathcal{L}^{\Omega}(\bm{\theta}) = \frac{1}{N_r}\sum \limits_{i=1}^{N_r} \left | u(\bm{x}_i;\bm{\theta}) - \widetilde{\mathbb{E}}\left[\tilde{q}(\Delta t;u(\cdot;\bm{\theta}), \bm{x}_i, \{\bm{B}_s\}_{0\leq s \leq \Delta t}) \middle | \bm{B}_0=\bm{x}_i\right]\right|^2,
\end{equation} 
while the training is in the same manner as the $q$-martingale loss.

% SECTION3
\section{Derivative-free network training for multiscale problems} \label{sec:proposed_method}
We are interested in a neural network-based approach to solving multiscale problems with no need for a well-designed network architecture nor direct reference to the homogenization theory \cite{BLP}. From the intrinsic averaging nature of representing the solution, the DFLM becomes a natural approach to approximate the homogenized solution of a multiscale problem. To extend the standard DFLM to multiscale problems, we investigate the effect of time stepping in solving the Brownian motion and its corresponding integrals. As in the time integration of multiscale problems \cite{ODEHMM, lee2013variable}, it turns out that micro- and macro-time steps play an important role in capturing the effective behavior of the multiscale solution. The micro-time step must be sufficiently small to resolve the fast variation of multiscale features, while the macro-time step must be sufficiently long enough to capture the effective averaged behaviors of the solution. For periodic problems, we show that the ratio between the two time steps is independent of the small periodicity. Therefore, the computational cost of the proposed method is independent of the small scale without using the homogenized coefficient.

We believe that the proposed method can be applied to a wide range of problems in the form of Eq.~\eqref{eq:Quasi-linear elliptic}, including parabolic problems. In the current study, we focus on the following elliptic boundary value problem
\begin{align}\label{eq:multiscaleElliptic}
\begin{split}
-\nabla \cdot \left(a^{\epsilon}\left(\bm{x}, u(\bm{x})\right)\nabla u(\bm{x})\right) &= f(\bm{x}) ~\textrm{in}~ \Omega \subset \mathbb{R}^d, \\
u(\bm{x}) &= g(\bm{x}) ~\textrm{on}~ \partial \Omega.
\end{split}
\end{align}
Here $f(\bm{x}) \in \mathbb{R}$ is the force term, and $g(\bm{x})\in \mathbb{R}$ is the Dirichlet boundary value. The permeability or conductivity coefficient $a^{\epsilon}(\bm{x}, u(\bm{x}))\in \mathbb{R}$ is the source of multiscale characteristics with a parameter $\epsilon\ll 1$ that represents the smallest scale involved in the problem. The variable coefficient $a^{\epsilon}$ can depend on $u(\bm{x})$, which yields a nonlinear multiscale problem. 
For the wellposedness of the PDE, the variable coefficient is uniformly bounded below
by a positive constant, that is, $a^{\epsilon} \geq a_{\min}>0$. We also assume that the coefficient is continuously differentiable with respect to $\bm{x}\in \Omega$. From the regularity of $a^{\epsilon}$, Eq.~\eqref{eq:multiscaleElliptic} can be formulated in the form of  Eq.~\eqref{eq:Quasi-linear elliptic} where $\bm{V}=\frac{\nabla_{\bm{x}} a^{\epsilon}}{2a^{\epsilon}}$ and $G=\frac{-f}{2a^{\epsilon}}$.

\subsection{micro- and macro-time stepping}
The application of the DFLM to Eq.~\eqref{eq:multiscaleElliptic} can be achieved by training a network to satisfy a corresponding martingale property, either the $q$-martingale (Eq.~\eqref{eq:stochastic_process}) or the $\tilde{q}$-martingale (Eq.~\eqref{eq:Brownian_martingale}). To avoid the time step restriction imposed by the drift term in solving Eq.~\eqref{eq:stochastic_process} in the $q$-martingale formulation, we use the $\tilde{q}$-martingale formulation described by the standard Brownian motion with no drift term
\begin{align}\label{eq:martingale_for_variable_coefficient_poisson_general}
\begin{split}
u(\bm{x}) &= \mathbb{E}\left[\tilde{q}(\Delta t;u, \bm{x}, \{\bm{B}_s\}_{0\leq s \leq \Delta t}) \middle | \bm{B}_0 = \bm{x} \right] \\
&= \mathbb{E}\left[\left(u(\bm{B}_{\Delta t}) - \mathcal{R}\left(\frac{-f}{2a^{\epsilon}},u,\Delta t\right)\right)\mathcal{D}\left(\frac{\nabla_{\bm{x}} a^{\epsilon}}{2a^{\epsilon}},u,\Delta t \right) \bigg | \bm{B}_0 = \bm{x}  \right].
\end{split}
\end{align}
Here $\mathcal{R}\left(\frac{-f}{2a^{\epsilon}},u,\Delta t\right)$ and $\mathcal{D}\left(\frac{\nabla_{\bm{x}} a^{\epsilon}}{2a^{\epsilon}},u,\Delta t \right)$ reads as in Eq.~\eqref{eq:Brownian_martingale}. From the learning perspective, the expectation term (i.e., RHS of Eq.~\eqref{eq:martingale_for_variable_coefficient_poisson_general}) is the target value of a neural network at $\bm{x}$, which is iteratively updated during the training procedure. 
By following the $\tilde{q}$-martingale formulation, the target random variable $\tilde{q}(\Delta t;u, \bm{x}, \{\bm{B}_s\}_{0\leq s \leq \Delta t})$ involves the multiscale characteristic through the integrals involving the multiscale coefficient $a^{\epsilon}$
\begin{equation}\label{eq:stochastic_integrals}
\int_0^{\Delta t}h^{\epsilon}(\bm{B}_s)ds ~~(h^{\epsilon}:\Omega \mapsto \mathbb{R}), \mbox{ and }
\int_0^{\Delta t}\bm{h}^{\epsilon}(\bm{B}_s)\cdot d\bm{B}_s ~~(\bm{h}^{\epsilon}:\Omega \mapsto \mathbb{R}^{d}),
\end{equation}
where $h^{\epsilon}=\frac{-f}{2a^{\epsilon}}$ or $\left|\frac{\nabla a^{\epsilon}}{2a^{\epsilon}}\right|^2$ and $\bm{h}^{\epsilon}=\frac{\nabla a^{\epsilon}}{2a^{\epsilon}}$. We can interpret
these integrals as the collected information from the Brownian walkers during the time period of $\Delta t$ in the multiscale media.

We introduce a micro-time step $\delta t$ for the discrete Brownian motion, which is simulated by the Euler-Maruyama method as 
\begin{equation}\label{eq:stoint-approx}
\bm{B}_{k\delta t}=\bm{B}_{(k-1)\delta t} + \sqrt{\delta t} \bm{Z}, \quad\bm{B}_0=\bm{x}, \quad\bm{Z}\sim \mathcal{N}\left(\bm{0}, \mathbb{I}_d\right),\quad k\in \mathbb{N}.
\end{equation}
The micro-time step must be shorter than or equal to the macro-time step $\Delta t$, the time period for the stochastic integration. 
For the micro- and macro-time steps satisfying $\Delta t = K \delta t$ (i.e., $K\in\mathbb{N}$ is the number of micro-time steps for one macro-time step), we estimate the stochastic integrals of Eq.~\eqref{eq:stochastic_integrals} as
\begin{equation}\label{eq:approximation-stochasitc-integral}
\begin{split}
&\int_0^{\Delta t}h^{\epsilon}(\bm{B}_s)ds \simeq \sum\limits_{k=0}^{K-1}h^{\epsilon}(\bm{B}_{k\delta t})\delta t,\\
&\int_0^{\Delta t}\bm{h}^{\epsilon}(\bm{B}_s)\cdot d\bm{B}_s \simeq \sum \limits_{k=1}^{K}\bm{h}^{\epsilon}\left(\bm{B}_{(k-1)\delta t}\right)\cdot \left(\bm{B}_{k\delta t}-\bm{B}_{(k-1)\delta t}\right).
\end{split}
\end{equation}
The micro-time step $\delta t$ must be chosen for the stability and accuracy of the numerical integration of the Brownian walkers, while the macro-time step $\Delta t$ determines the size of the sample's neighborhood at a given $\bm{x}$. We note that the size of the neighborhood is related to the expectation of the target variable. 
For instance, a macro-time step longer than the exit time $\tau=\inf \{s: \bm{B}_s \notin \Omega \}$ provides the exact solution (i.e., Feynman-Kac formula) as the target value, but it may require an excessive number of micro-time steps for the Brownian motion to reach the boundary. On the other hand, if the macro-time step is too small, such as $\Delta t = \delta t$, the target will reflect only small neighborhood information limited to learning the over-the-domain features of the solution.
In the original work of DFLM \cite{han2020derivative}, the micro- and macro-time steps are set to be equal, $\Delta t = \delta t$, which fails to solve multiscale problems (see Fig.~\ref{fig:macrotimestep}  for a failure of the case $\delta t=\Delta t$ for a multiscale problem).

\subsection{Bounds of micro- and macro-time steps}
The ratio between the micro- and macro-time steps affects the computational cost of the proposed method. For accuracy and stability, a small micro-time step is preferred while we need a large macro-time step to have an appropriate averaging effect. In this section, we provide the upper and lower bounds of the micro- and macro-time steps, respectively, for a periodic multiscale coefficient $a^{\epsilon}(\bm{x})=a(\frac{\bm{x}}{\epsilon})$ where $a(\cdot)$ is 1-periodic in all directions. The result shows that the ratio between the two time steps is independent of $\epsilon$ and thus the computational cost of the proposed method remains constant.

\begin{figure}[t]
\centering
\includegraphics[width=1\textwidth]{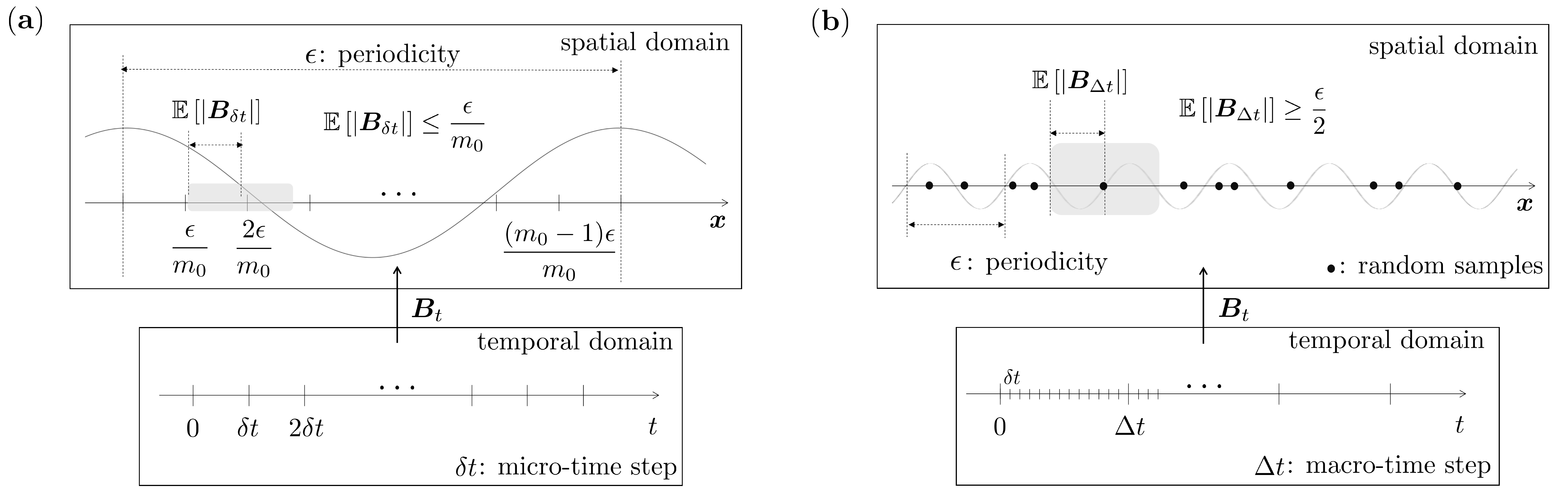}
\caption{Schematic diagrams for choosing (a) micro-time step $\delta t$ and (b) macro-time step $\Delta t$.} 
\label{fig:schematic_diagrams}
\end{figure} 
In the evaluation of the stochastic integrals Eq.~\eqref{eq:approximation-stochasitc-integral}, the Brownian walker must be sampled enough so that the behaviors of the function $h^{\epsilon}(\bm{x})$ and $\bm{h}^{\epsilon}(\bm{x})$ are properly reflected. As we follow the $\tilde{q}$-martingale formulation, which involves the standard Brownian walker with no drift term, the micro-time step is mainly restricted by the accuracy. 
We require that the average distance between successive Brownian walkers $\bm{B}_{k\delta t}$ and $\bm{B}_{(k+1)\delta t}$ to be less than $\frac{\epsilon}{m_0}$ for a parameter $m_0\in\mathbb{N}$, so that we can resolve the microscale variation of the integrands. That is, a large $m_0$ is required to increase the accuracy of the stochastic integral approximation \eqref{eq:stoint-approx} (see a schematic diagram of the micro-time stepping in Fig.~\ref{fig:schematic_diagrams}-(a)). 
From the property of the Brownian motion, we have 
\begin{equation}\label{eq:microstep_upperbound_derivation}
\frac{\epsilon}{m_0} \geq \mathbb{E}[|\bm{B}_{(k+1)\delta t}-\bm{B}_{k\delta t}|] = \kappa(d)\sqrt{\delta t},
\end{equation}
where $\kappa(d)$ is a constant depends only on the dimension $d$. Thus, the upper bound of the micro-time step is given by
\begin{equation}\label{eq:microstep_upperbound}
\delta t \leq \frac{1}{\kappa^2(d)}\left(\frac{\epsilon}{m_0} \right)^2,
\end{equation}
which we denote as $\overline{\delta t} (m_0;\epsilon)$.
For instance,  the upper bound of the micro-time step for the 2-dimensional case is $\overline{\delta t}(m_0;\epsilon)=\frac{2}{\pi}\left(\frac{\epsilon}{m_0} \right)^2$ as $\kappa(2)=\sqrt{\frac{\pi}{2}}$.

\begin{figure}[t]
\centering
\includegraphics[width=1\textwidth]{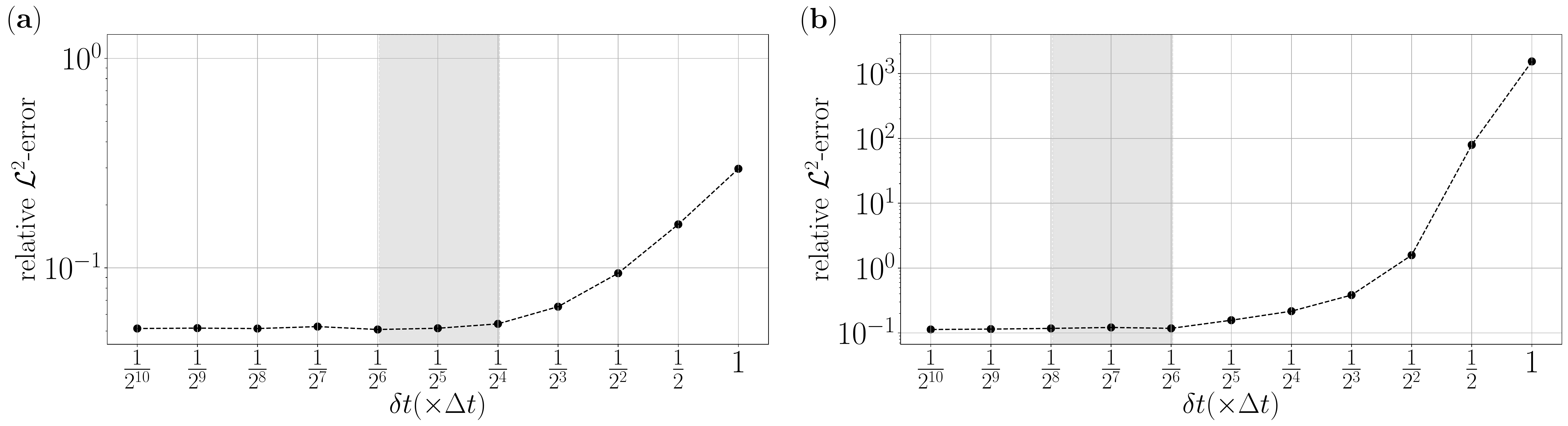}
\caption{The accuracy of neural network approximations for the multiscale test  problem  with various micro-time steps under fixed macro-time step. The relative $\mathcal{L}^2$-error is averaged over $20$ independent trials.
(a): $\epsilon=0.1$, $\Delta t=1\times10^{-3}$, (b): $\epsilon=0.05$, $\Delta t=1\times10^{-3}$. The dark regions in both plots represent the range $[\overline{\delta t}(20;\epsilon),\overline{\delta t}(10;\epsilon)]$ as reference.} 
\label{fig:microtimestep}
\end{figure}

To check the validity of the upper bound, we test various micro-time steps for two different $\epsilon$ values for a test problem (the test model is \eqref{eq:multiscaleElliptic} in $\Omega=[0,1]^2$ with $a^{\epsilon}(\bm{x})=2+\sin\left(\frac{2\pi}{\epsilon}x_1\right)\cos\left(\frac{2\pi}{\epsilon}x_2\right)$, $f(\bm{x})=10$, and $g(\bm{x})=0$ following the other test setup in section \ref{sec:Numerical Experiments}). Except the micro-time steps and $\epsilon$, all other values are fixed (the same training parameters in section \ref{subsec:linearProblem_periodic} except the learning rate parameters $(\alpha_0,\gamma)=(10^{-4},0.9)$). The relative $\mathcal{L}^2$-errors as a function of micro-time step sizes are shown in Fig.~\ref{fig:microtimestep} ((a) for $\epsilon=1\times 10^{-1}$ and (b) for $5\times 10^{-2}$, while the macro-time step is fixed at $1\times 10^{-3}$ in both cases). The horizontal axis is the ratio between the micro- and macro-time steps, ranging from $\frac{1}{2^{10}}$ to $1$. When the ratio is close to 1, the relative error is dominated by the error related to the micro-time step. As we decrease the ratio, both lines show stabilized error behaviors after the ratios $\frac{1}{2^6}$ and $\frac{1}{2^4}$ for $\epsilon=5\times 10^{-2}$ and $1\times 10^{-1}$, respectively. These values correspond to the upper bound estimates when $m_0=10$.  

\subsubsection{A lower bound of macro-time step}\label{subsec:lower_bound_of_macro-time step}
The length of the macro-time step $\Delta t$ determines the size of the neighborhood covered by the Brownian walker to estimate the target value (i.e., the homogenized solution). If the macro-time step is too short, the walker will see only a small fraction of the required information consisting of sporadic pieces of the periodic neighborhood.

For a single training sample point $\bm{x}$, the $\epsilon$-length neighborhood, $\mathcal{R}_{\epsilon}(\bm{x}):=\bigl\{\bm{z}:\|\bm{z} \allowbreak -\bm{x}\|_{\infty}\leq \frac{\epsilon}{2}\bigr\}$, contains all information about physical or medium properties written in the $\epsilon$-periodic coefficient. Thus, the requirement for the macro-time step is to guarantee that the Brownian walker passes through the neighborhood $\mathcal{R}_{\epsilon}(\bm{x})$
\begin{equation}
\frac{\sqrt{d}\epsilon}{2}=(\text{circumradius of } \mathcal{R}_{\epsilon}(\bm{x}))   \leq \mathbb{E}[|\bm{B}_{\Delta t}|]  = \kappa(d)\sqrt{\Delta t},
\end{equation}
where $\kappa(d)$ is the same constant in Eq.~\eqref{eq:microstep_upperbound_derivation}. This yields the lower bound of the macro-time step
\begin{equation}\label{eq:macrolowerbound}
\frac{d\epsilon^2}{4\kappa^2(d)} \leq \Delta t,
\end{equation}
which we denote as $\underline{\Delta t}^{\ast}(\epsilon)$ (see Fig.~\ref{fig:schematic_diagrams}-(b) for schematic diagram). The lower bound for the 2-dimensional case, for example, is given by $\underline{\Delta t}^{\ast}(\epsilon)=\frac{\epsilon^2}{\pi}$ as $\kappa(2)=\sqrt{\frac{\pi}{2}}$. We note that the ratio between the macro- and micro-time stpes is independent of $\epsilon$
\begin{equation}\label{eq:micro-macro-ratio}
\frac{\underline{\Delta t}^{\ast}(\epsilon)}{\overline{\delta t}(m_0)}~=~\frac{dm_0^2}{4}~:=~K_0(m_0),
\end{equation}
which depends on the discretization parameter $m_0$ related to the accuracy of the stochastic integrals \eqref{eq:approximation-stochasitc-integral}. Therefore, the computational cost of the proposed method is independent of the small-scale parameter $\epsilon$.

\begin{figure}[t]
\centering
\includegraphics[width=1\textwidth]{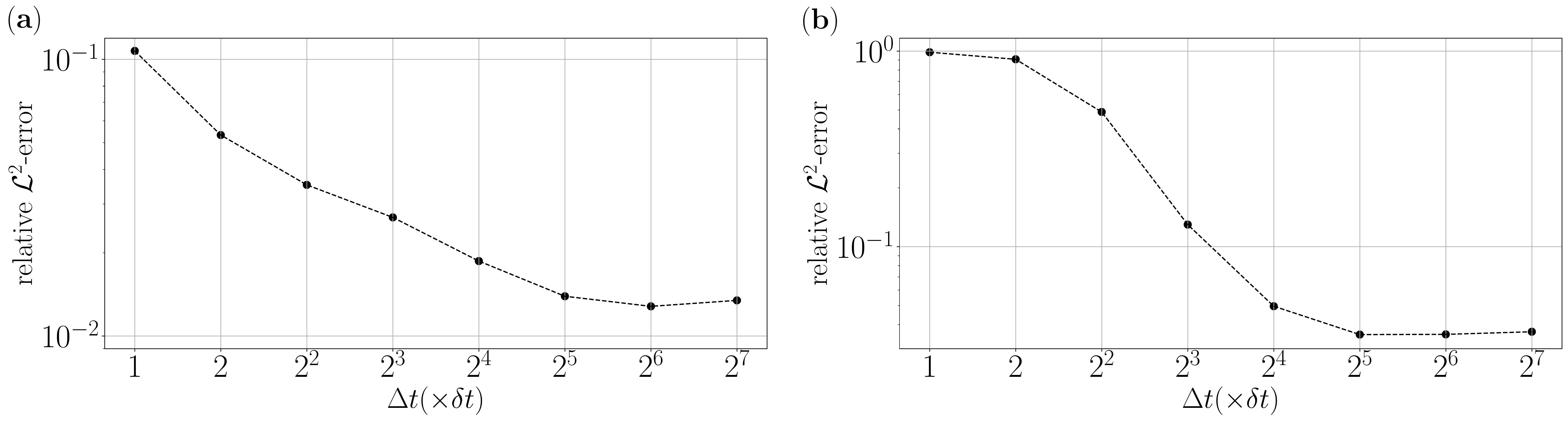}
\caption{The accuracy of neural network approximations for the multiscale test  problem with various macro-time steps under the choice of micro-time step referred to the upper bound $\overline{\delta t}(12,\epsilon)$. The relative $\mathcal{L}^2$-error is averaged over $20$ independent trials.
(a): $\epsilon=5\times 10^{-2}$, (b): $\epsilon=5 \times 10^{-3}$.} 
\label{fig:macrotimestep}
\end{figure}

Instead of validating the lower bound of the macro-time step directly, we show that the ratio between the two time steps remains constant. Fig.~\ref{fig:macrotimestep} shows the relative errors for $\epsilon=5\times 10^{-3}$ and $5\times 10^{-2}$, respectively, applied to the same test for the micro-time step validation. 
We choose the micro-time step $\delta t$ in reference to the upper bound (i.e. $\delta t \simeq \overline{\delta t}(m_0,\epsilon)$ with $m_0=12$) so that the error from the micro-time stepping remains comparable between the two $\epsilon$ values. As the ratio increases, the relative errors decrease and they become stable when the ratio becomes larger than $2^5$ for both $\epsilon$ values, which is close to the estimate for the ratio $K_0(12)=\frac{2\times 12^2}{4}=72$ (note that the test problem is a 2-dimensional problem and thus $d=2$).

% SECTION4
\section{Numerical Experiments}\label{sec:Numerical Experiments}
In this section, we validate the robustness and effectiveness of the proposed derivative-free loss formulation through a suite of multiscale test problems. In particular, we use the interior loss function corresponding to $\tilde{q}$-martingale in Eq.~\eqref{eq:loss_interior_continuous_ver2} driven by the standard Brownian motion, along with the boundary loss term $\mathcal{L}^{\partial \Omega}(\bm{\theta})$ in Eq.~\eqref{eq:loss_boundary_dirichlet}.
In all experiments, we use the standard multilayer perceptrons (MLPs) with the ReLU activation function. We train neural networks via stochastic gradient descent (SGD) using the Adam optimizer \cite{kingma2014adam} with learning parameters $\beta_1=0.99$ and $\beta_2=0.99$, and all other trainable parameters are initialized from the Glorot normal distribution \cite{glorot2010understanding}. We employ an exponential decay learning rate with an initial rate $\alpha_0$ and decay rate $\gamma$ per $1000$ training iterations. The specific values for $\alpha_0$ and $\gamma$ will be specified in each test problem. In the derivative-free loss formulation, the locations of stochastic walkers at each iteration become the random training samples.
In this work, we set the locations of the walkers from the uniform distribution in the domain to control the effect of sampling methods on the learning procedure in which the walkers are resampled at each training process. We obtain reference solutions $u$ using the FEM method with sufficiently fine mesh sizes after testing convergence. The accuracy of a neural network solution $\tilde{u}$ is measured with the relative $\mathcal{L}^2$-error, $\frac{\|\tilde{u}-u\|_{2,\Omega}}{\|u\|_{2,\Omega}}$
where $\mathcal{L}^2$-norm is computed on the equidistant $501\times 501$ grid points.

\subsection{Linear multiscale problem}\label{sec:linear_multiscale_periodic}\label{subsec:linearProblem_periodic}
\begin{figure}[!htbp]
\centering
\includegraphics[width=0.89\textwidth]{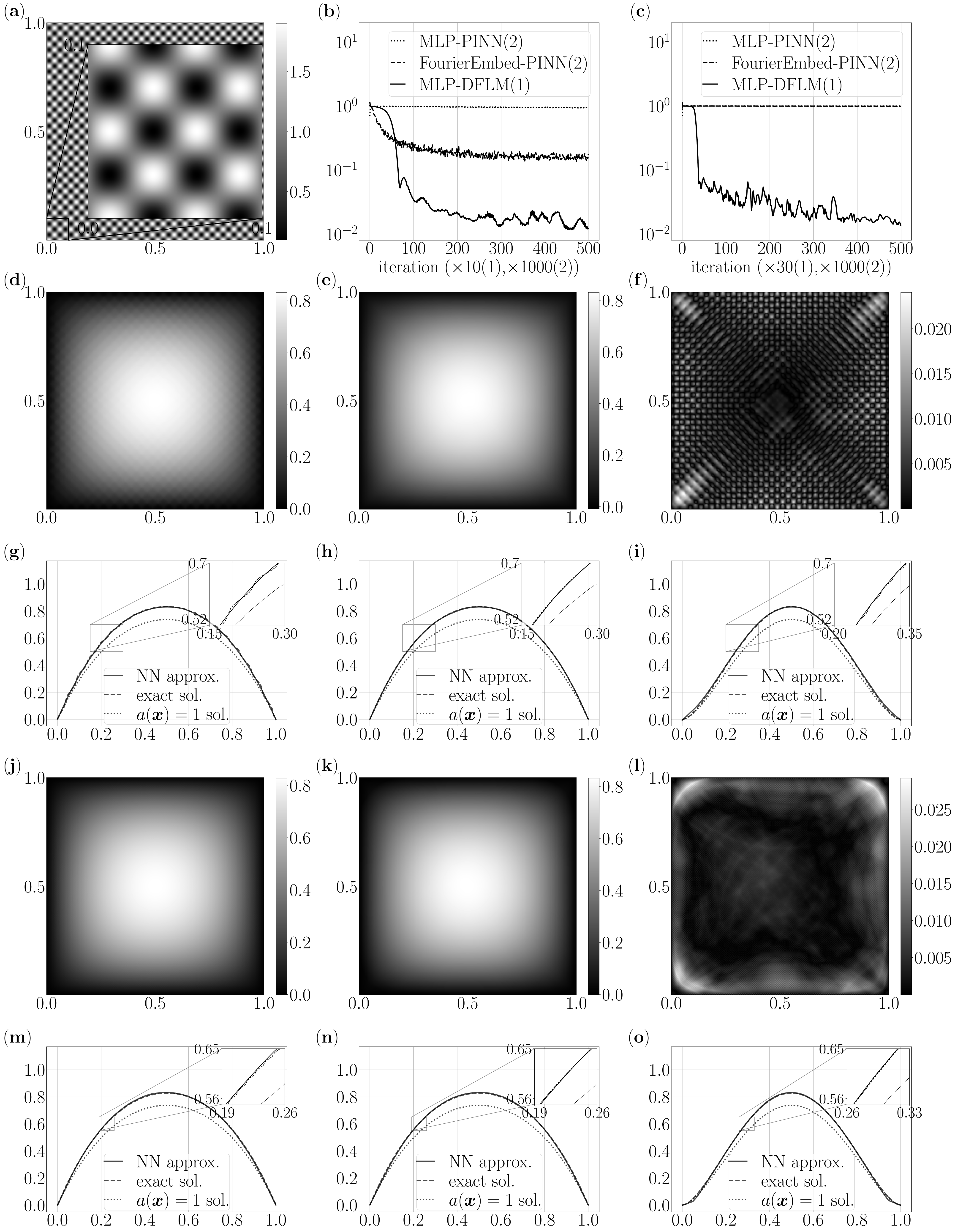}
\caption{Linear Poisson equation with the periodic coefficient. (a): the distribution of coefficient $\alpha^{\epsilon}(\bm{x})$ (in Eq.~\eqref{eq:linearPoisson_coef} or \eqref{eq:nonlinearPoisson_coef}) for $\epsilon=0.05$.  The following pairs read as ``for $\epsilon=0.05$ and $\epsilon=0.01$,  respectively"; (b),(c): training procedure (relative $\mathcal{L}^2$-error),  (d),(j): FEM reference solution,  (e),(k): MLP approximation, (f),(l): pointwise error of approximation,  (g),(m): horizontal cross-section ($x_2=0.5$) of approximation, (h),(n): vertical cross-section ($x_1=0.5$) of approximation, (i),(o): diagonal cross-section ($x_1=x_2$) of approximation.} 
\label{fig:linearPoisson2D_approximation}
\end{figure}

The first example is the linear Poisson problem Eq.~\eqref{eq:multiscaleElliptic} in the unit square $\Omega=[0,1]^2$ with a periodic coefficient 
\begin{equation}
\label{eq:linearPoisson_coef}
a^{\epsilon}(\bm{x},u(\bm{x}))=\alpha^{\epsilon}(\bm{x})=1+0.9\sin\left(\frac{2\pi}{\epsilon}x_1\right)\cos\left(\frac{2\pi}{\epsilon}x_2\right).
\end{equation}
To impose moderate and strong multiscale characteristics, we use $\epsilon=0.05$ and $0.01$, respectively (the plots of $\alpha^{\epsilon}(\bm{x})$ for $\epsilon=0.05$ and $0.01$ are shown in Fig.~\ref{fig:linearPoisson2D_approximation}-(a) and 
Fig.~\ref{fig:nonlinearPoisson2D_approximation}-(a), respectively). The boundary value is homogeneous with $g(\bm{x})=0$. To have a non-trivial solution, we have a constant $f(\bm{x})=10$.
%
%
% Begin Figure

The MLP network has four hidden layers of dimension $200$, and we use $N_r=400$ samples for the interior, $N_s=300$ samples for target estimation, and $N_b=400$ for the boundary in both $\epsilon$ values.
We train the MLP neural network $u(\bm{x};\bm{\theta})$ to satisfy the $\tilde{q}$-martingale property Eq.~\eqref{eq:martingale_for_variable_coefficient_poisson_general} where the reward $\mathcal{R}$ and the discount $\mathcal{D}$ do not depend on $u(\cdot;\bm{\theta})$. We select an effective micro- and macro-time steps, $\delta t$ and $\Delta t$, based on the proposed bounds. We choose the micro-time step according to the upper bound in Eq.~\eqref{eq:microstep_upperbound}; $\delta t = 1.10\times 10^{-5} \simeq \overline{\delta t}(12;0.05)$ for $\epsilon=0.05$ and $\delta t = 4.42\times 10^{-7} \simeq \overline{\delta t}(12;0.01)$ for $\epsilon=0.01$. Then, we set the macro-time step $\Delta t = K \delta t$ where the number of micro-stepping is chosen as $K=72$ in reference to $K_0(m_0)$ in Eq.~\eqref{eq:micro-macro-ratio}. 
We emphasize that the choice of $K$ does not depend on the $\epsilon$ scale.  
Fig.~\ref{fig:linearPoisson2D_approximation} summarizes training procedure ((b) for $\epsilon=0.05$ and (c) for $\epsilon=0.01$) of neural networks during $5000$($\epsilon=0.05$) and $15000$($\epsilon=0.01$) iterations using learning rate parameters  $(\alpha_0, \gamma)=( 10^{-4}, 0.85)$ for $\epsilon=0.05$ and $(\alpha_0, \gamma)=(7\times10^{-5}, 0.85)$ for $\epsilon=0.01$, and detail approximation results ((d)-(i) for $\epsilon=0.05$ and (j)-(o) for $\epsilon=0.01$). The neural networks approximate the global characteristics of the solutions as homogenized correspondence (solid curves in Fig.~\ref{fig:linearPoisson2D_approximation} (g)-(i) for $\epsilon=0.05$, (m)-(o) for $\epsilon=0.01$), each of which has relative $\mathcal{L}^2$-errors $1.21\times 10^{-2}$ for $\epsilon=0.05$ and $1.38\times 10^{-2}$ for $\epsilon=0.01$. In particular, the behavior of the approximation is far different from the solution obtained by replacing the highly oscillatory coefficient with its average value over the domain, $a(\bm{x})=\frac{1}{|\Omega|}\int_{\Omega}\alpha^{\epsilon}(\bm{x})d\bm{x}=1$ (dotted curves in  Fig~\ref{fig:linearPoisson2D_approximation}. (g)-(i) for $\epsilon=0.05$ and (m)-(o) for $\epsilon=0.01$), which shows that the microscale feature makes non-trivial effects on the macroscopic behavior of the solution.

We also solve the problem using the PINN method \cite{raissi2019PINN} for comparison. We use standard MLPs and the Fourier feature embedded neural networks  \cite{wang2021eigenvector}. Moreover, we employ the adaptive weights algorithm \cite{wang2022and}, which adjusts the discrepancy in convergence rates of interior and boundary loss components during the training procedure. While we have made our best effort to train by varying dimensions of neural networks, activation functions, sample size, and Fourier feature embedding parameters (i.e., standard deviation $\sigma$ for random wave numbers driven from $\mathcal{N}(0,\sigma^2)$), we could not achieve an accurate approximation with slow convergence. We present the training procedures ($5\times 10^{5}$ iterations) using the standard MLP (four hidden layers of dimension $200$, $\tanh$ activation) and  Fourier feature embedded neural network (200-dimensional embedding with parameter $\sigma=30$ followed by MLP with four hidden layers of dimension $200$ and $\tanh$ activation) using the training samples of size $N_r^{\text{PINN}}=12000$ (interior sample), $N_b^{\text{PINN}}=400$ (boundary sample) in Fig.~\ref{fig:linearPoisson2D_approximation}-(b) and (c). In comparison to the PINN method, the proposed method requires relatively small numbers of gradient descent steps (epochs) to achieve the approximation comparable to the reference solution. The proposed method requires $(N_r\times N_s\times K)$ samples for the target estimation and $N_r$ samples for $\bm{\theta}$-derivative computation of the neural network at each iteration. On the other hand, the PINN method requires $N_r^{\text{PINN}}$ samples for both $\bm{x}$- and $\bm{\theta}$-derivative computations of the neural network. The proposed method saves cost with a small number of iterations achieving an accurate approximation to the macroscopic behavior of the multiscale solution.

\subsection{Nonlinear multiscale problem}
The second example is a nonlinear Poisson equation Eq.~\eqref{eq:multiscaleElliptic} in $\Omega=[0,1]^{2}$ with the coefficient that depends on $u(\bm{x})$
\begin{equation}\label{eq:nonlinearPoisson_coef}
a^{\epsilon}(\bm{x},u(\bm{x}))=1+\alpha^{\epsilon}(\bm{x})u^{2}(\bm{x})~~~\textrm{where}~ \alpha^{\epsilon}(\bm{x})=1+0.9\sin\left(\frac{2\pi}{\epsilon}x_1\right)\cos\left(\frac{2\pi}{\epsilon}x_2\right).
\end{equation}
As in the linear case, we test $\epsilon=0.05$ and $0.01$ and the homogeneous boundary value $g(\bm{x})=0$, while the force term $f(\bm{x})$ is set to $50$.
For this nonlinear problem, the $\tilde{q}$-martingale property for training the neural network $u(\bm{x};\bm{\theta})$ is Eq.~\eqref{eq:martingale_for_variable_coefficient_poisson_general} with the reward $\mathcal{R}$ and the discount $\mathcal{D}$ factors 
\begin{equation}
\mathcal{R}\left(\frac{-f}{2(1+\alpha^{\epsilon}u^2)},u,\Delta t\right),~~~~ \mathcal{D}\left(\frac{u^2\nabla \alpha^{\epsilon} + 2\alpha^{\epsilon}u\nabla u}{2(1+\alpha^{\epsilon}u^2)},u,\Delta t \right).
\end{equation}

%We note that the reward and the discount factor depend on the neural network $u(\bm{x};\bm{\theta})$. The dependency of varying neural network might cause the uncertainty to choose the time steps for estimating stochastic integrals since we have no direct information about the length scale of the neural network. Fortunately, according to the F-principle \cite{xu2019frequency}, the neural network $u(\bm{x};\bm{\theta})$ learn the low frequency first, that is, its length scale is longer than the scale of $\epsilon \ll 1$ during the training procedure. It follows that the integrands in the discount factor still has the finest scale $\epsilon$ and we can choose the effective micro-macro time steps according to the proposed bounds. \yl{Need to clean the above paragraph; explanation is not clear.}

\begin{figure}[!htbp]
\centering
\includegraphics[width=0.89\textwidth]{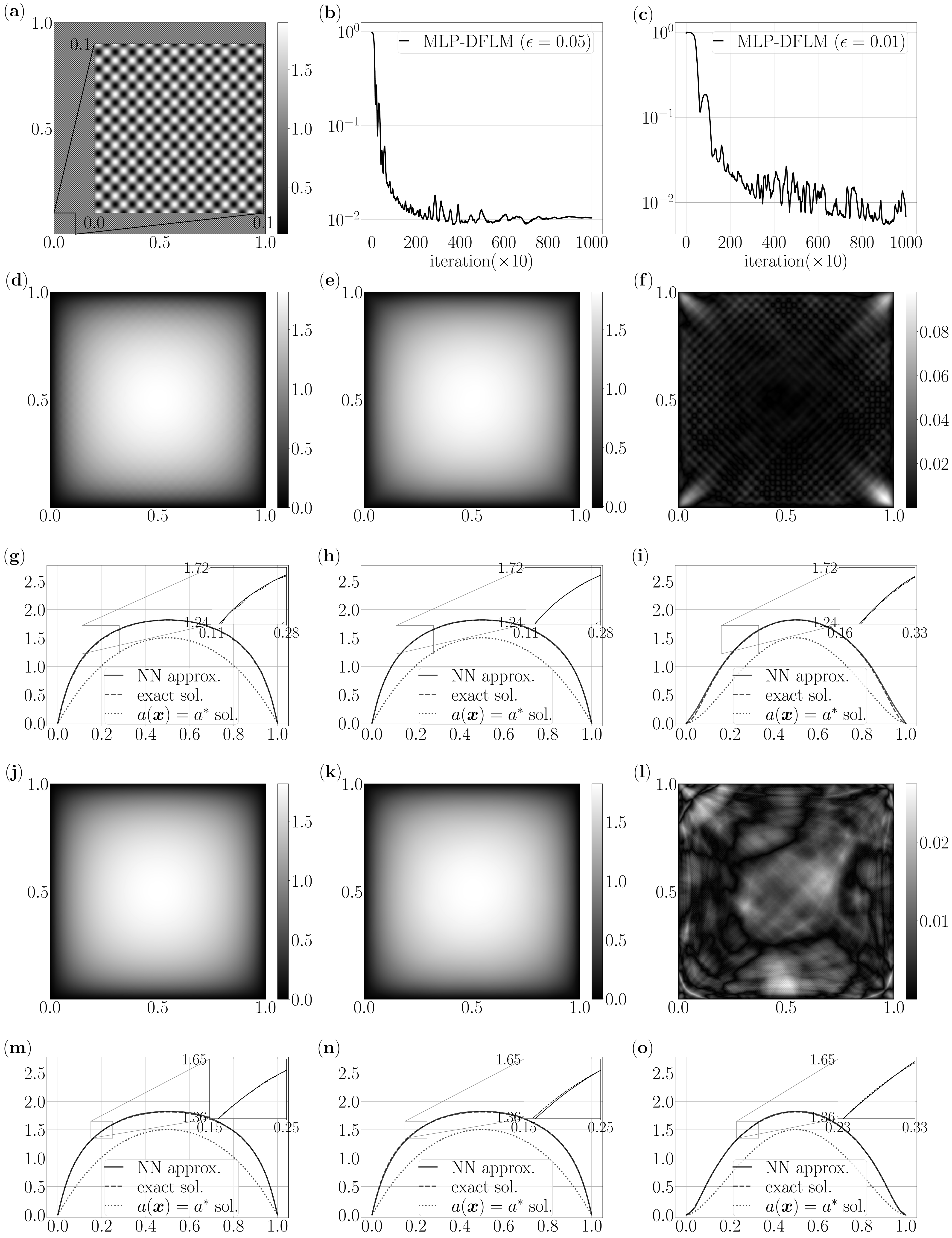}
\caption{Nonlinear Poisson equation with periodic coefficient. (a): the distribution of coefficient $\alpha^{\epsilon}(\bm{x})$ (in Eq.~\eqref{eq:linearPoisson_coef} or \eqref{eq:nonlinearPoisson_coef})  for $\epsilon=0.01$.  The following pairs read as ``for $\epsilon=0.05$ and $\epsilon=0.01$,  respectively"; (b),(c): training procedure (relative $\mathcal{L}^2$-error),  (d),(j): FEM reference solution,  (e),(k): MLP approximation, (f),(l): pointwise error of approximation,  (g),(m): horizontal cross-section ($x_2=0.5$) of approximation, (h),(n): vertical cross-section ($x_1=0.5$) of approximation, (i),(o): diagonal cross-section ($x_1=x_2$) of approximation.} 
\label{fig:nonlinearPoisson2D_approximation}
\end{figure}

Regarding the time steps, we use the same values as in the linear case ($(\delta t,\Delta t)=(1.10\times 10^{-5},7.92\times 10^{-4})$ for $\epsilon=0.05$ and $(4.42\times 10^{-7},3.18\times 10^{-5})$ for $\epsilon=0.01$), and train the standard MLP with four hidden layers of dimension $200$ using the sample size $N_r=600$, $N_s=400$, and $N_b=400$ in both $\epsilon$ values. We present the training procedures (learning rate parameters $(\alpha_0,\gamma)=(5\times 10^{-4},0.5)$ for $\epsilon=0.05$, $(9\times 10^{-4},0.8)$ for $\epsilon=0.01$) and approximation results during $10^4$ iterations in Fig.~\ref{fig:nonlinearPoisson2D_approximation}. The proposed method captures the macroscopic behavior of the solution closed to the homogenized solution (approximation over the domain in Fig.~\ref{fig:nonlinearPoisson2D_approximation}. (d)-(f) for $\epsilon=0.05$ and (j)-(l) for $\epsilon=0.01$ and the cross-sections in Fig.~\ref{fig:nonlinearPoisson2D_approximation} (g)-(i) for $\epsilon=0.05$ (m)-(o) for $\epsilon=0.01$) with relative $\mathcal{L}^2$-errors $1.04\times 10^{-2}$ for $\epsilon=0.05$ and $6.9\times 10^{-3}$ for $\epsilon=0.01$. We also compare the approximations from the solutions of linear Poisson equation with constant coefficients  approximated by empirical averaged value over the domain  $a(\bm{x})=a^{\ast}=\frac{1}{|\Omega|}\int_{\Omega}\left(1+\alpha^{\epsilon}(\bm{x})u^2(\bm{x};\bm{\theta}^{\ast})\right)d\bm{x}$ ($u(\cdot;\bm{\theta}^{\ast})$ neural network approximations). The cross-sections in  Fig.~\ref{fig:nonlinearPoisson2D_approximation} show that the homogenized coefficient corresponding to the neural network approximation is different from the heuristic average values.

In both the linear and the nonlinear examples, the proposed method approximates the corresponding homogenized solutions with low relative $\mathcal{L}^2$-errors less than or comparable to 1\%. The pointwise errors of approximations (Fig.~\ref{fig:linearPoisson2D_approximation}-(f),(l) and Fig.~\ref{fig:nonlinearPoisson2D_approximation}-(f),(l)) have relatively high pointwise errors near the four corners of the domain. We interpret that the estimation  of boundary-hitting locations of the Brownian samples (details in \cite{han2020derivative}) becomes less accurate near the corners and its effect on target estimation is relatively sensitive in a highly oscillatory environment. Possible directions for improvements along this direction include the usage of smooth approximation of the right-angle corners or adaptive sampling near the corners to moderate the negative effect. We leave this investigation as future work.

\subsection{Random-field coefficient multiscale problem}
\begin{figure}[h!]
\centering
\includegraphics[width=1.0\textwidth]{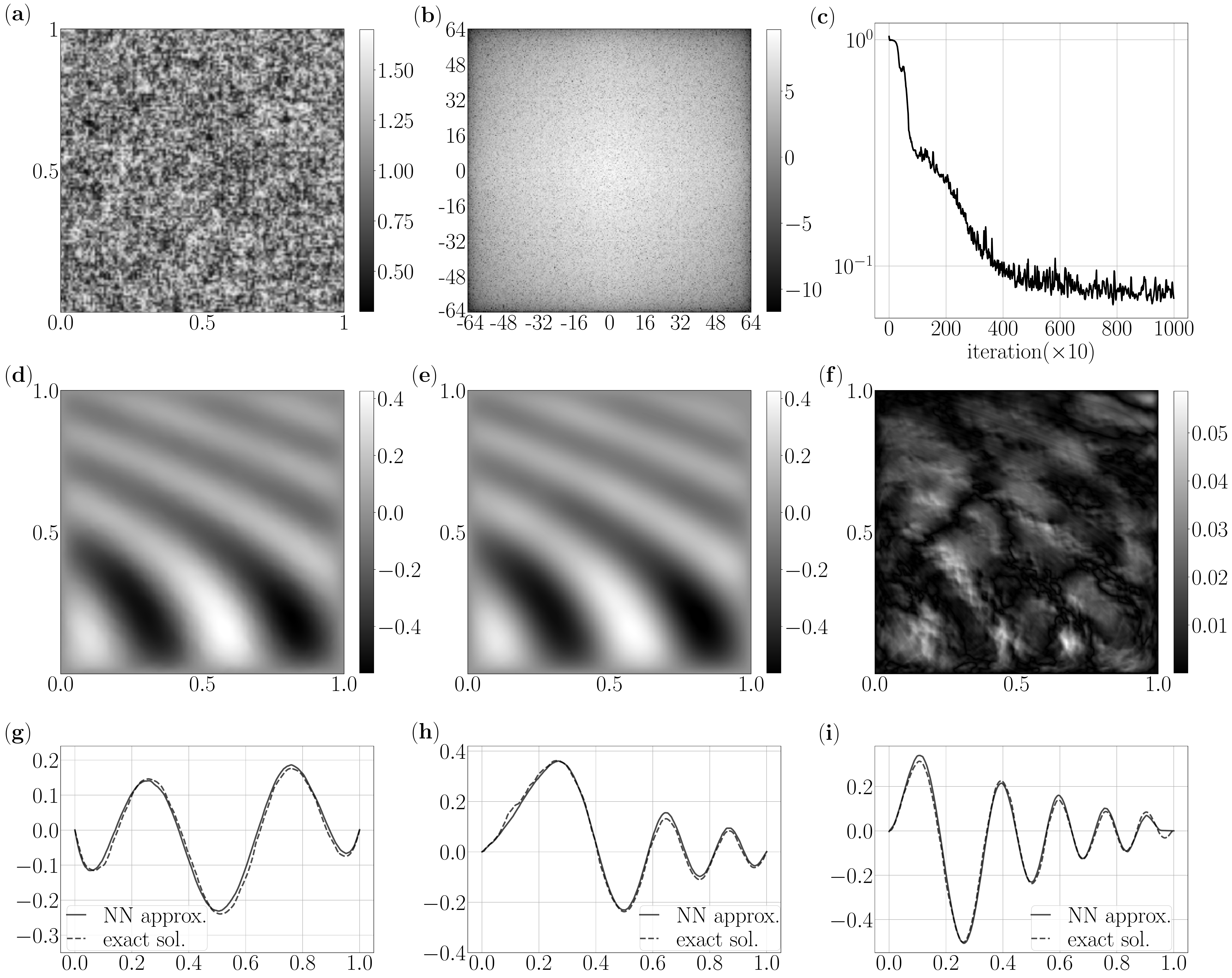}
\caption{Linear Poisson equation with random field coefficient. (a): the distribution of the coefficient in the physical domain, (b): the distribution of the coefficient in the Fourier domain (log-scaled amplitude), (c): training procedure (relative $\mathcal{L}^2$-error),  (d): FEM reference solution,  (e),: MLP approximation, (f): pointwise error of approximation,  (g): horizontal cross-section ($x_2=0.5$) of approximation, (h): vertical cross-section ($x_1=0.5$) of approximation, (i): diagonal cross-section ($x_1=x_2$) of approximation.} 
\label{fig:randFieldPoisson_approximation}
\end{figure}

The previous two examples have explicit scale separation represented by $\epsilon$. Our last example is a multiscale problem without scale separation. 
For the linear Poisson Eq.~\eqref{eq:multiscaleElliptic} in $\Omega=[0,1]$, we use a non-separable scale multiscale coefficient, which is a random field in $\Omega$ whose Fourier spectrum has a decay rate of $\mathcal{O}(k^{-1})$ (see Fig.~\ref{fig:randFieldPoisson_approximation}-(a) and -(b) for the plots of the random field in the physical and the Fourier domains). For a non-trivial solution and well-posedness, the force term $f(\bm{x})$ is set to
$100\sin(4\pi x_1+6\pi x_2^2)$, while the boundary value $g(\bm{x})=0$. The $\tilde{q}$-martingale property corresponding to Eq.~\eqref{eq:multiscaleElliptic} for training the neural network $u(\bm{x},\bm{\theta})$ is the same as Eq.~\eqref{eq:martingale_for_variable_coefficient_poisson_general} except $a^{\epsilon}$ read as $a^{\text{rand}}$.

As a measure to determine the micro- and macro-time steps without explicit scale separation, we use the spatial correlation of the random field coefficient. As the random field is isotropic (see Fig.~\ref{fig:randFieldPoisson_approximation}-(b) for the log-scale plot of the Fourier amplitudes), we use the average of the $x$- and $y$-directional spatial correlations, 0.0135, as $\epsilon$ to determine the time steps. The micro-time step $\delta t$ is set to $8.05\times 10^{-7}$ which corresponds to the upper bound in Eq.~\eqref{eq:microstep_upperbound} with $m_0=12$. For the macro-time step, we use the lower bound in Eq.~\eqref{eq:macrolowerbound} with a larger $\epsilon=2\times 0.0135$ so that the walkers can sufficiently explore the neighborhood, which yields $\Delta t=2.32\times 10^{-4}$.

We train the standard MLP with five hidden layers of dimension $300$ using the sample size $N_r=1600$,  $N_s=600$, and $N_b=400$. Fig.~\ref{fig:randFieldPoisson_approximation} summarizes the training procedure (Fig.~\ref{fig:randFieldPoisson_approximation}-(c)) and approximation results (Fig.~\ref{fig:randFieldPoisson_approximation}. (d)-(i)) after $10^4$ iterations with learning rate parameters $(\alpha_0,\gamma)=(5\times 10^{-4},0.8)$. The proposed approach captures the global macroscopic behavior of the solution with performance degradation compared to the previous two tests. The relative error is $7.21\times 10^{-2}$, which is 7 times larger than the previous tests. 

\section{Discussions and conclusions}\label{sec:discussion}
We proposed a neural network-based approach to solve multiscale problems. Due to non-trivial interactions between different scale components, resolving all relevant scales remains a challenge for multiscale problems. Our approach uses the derivative-free loss formulation (DFLM \cite{han2020derivative}) using an equivalent stochastic representation of a class of partial differential equations. The proposed method does not require the pre-computation of the homogenized coefficient or a particular network architecture.
The proposed method shows robust results for the standard periodic multiscale problem along with nonlinear and random field multiscale problems.

As the proposed method involves a stochastic formulation, two time steps are related to solving the stochastic system and the period to calculate averages. We showed that the ratio between the two time steps remains constant so that the computational cost of the proposed method is independent of the period for the periodic multiscale problem. For the random field problem, we used the spatial correlation to determine the time steps. This yields a result capturing the macroscopic behavior of the solution, but the performance degraded compared to periodic problems. We believe that the performance degradation comes from non-optimal time stepping. Although we have not studied in the current study, the number of samples can affect the performance of homogenization. The average size of the nearest neighborhood at each sample location is of $\mathcal{O}(N_r^{\frac{-1}{d}})$ \cite{bhattacharyya2008mean}. For non-separable scale problems, the time stepping must depend on the number of samples in addition to time stepping. Thus, it would be natural to investigate the effect of time stepping and the number of samples for non-separable scale problems.

As an approach to increasing the random walkers' averaging process, there are several methods to consider. In the multiscale time integration, a Kernel-based approach \cite{ODEHMM} has been proposed to handle non-matching periodicity. Such a method often requires specifying the slow and fast variables, which can be challenging to apply to PDEs as such specification is not straightforward. A variable time stepping \cite{lee2013variable} can also be considered as an alternative to improve the averaging accuracy without specifying the scale separation.
In addition to the length of the macro-time step, the number of samples and their distribution can also affect the averaging neighborhood of the walkers. In the current study, we used uniform resampling of the samples at each training iteration. We plan to investigate how the resampling distribution can affect the averaging neighborhood in capturing the homogenized solution. 

In the current study, we have focused on multiscale elliptic problems for the proposed method. The standard DFLM has been successfully applied to constant-coefficient elliptic and parabolic equations, and thus it is natural to use the proposed method for parabolic problems. Parabolic problems include actual time scale as time-dependent problems. The micro- and macro-time steps of the proposed method are instead related to the stochastic formulation as a local average than the actual time scale of a PDE model. We leave future work to speculate the actual time scale in solving multiscale problems using the derivative-free method for multiscale problems.

Our interest in multiscale problems is not limited to homogenized solutions. Recent work shows that a hierarchical design of a network can expedite the training process of learning multiscale problems \cite{han2021hierarchical}. The rationale for the success of the hierarchical learning lies in capturing different scale components with their corresponding network structures; for example, Fourier features embedding networks with disparate characteristic scales. We believe that our proposed method can serve as the first-level approximation in hierarchical learning. At the same time, the microscale details can be effectively captured by a particular network designed to capture only high wavenumber components. We are investigating the applicability of the proposed method in the context of hierarchical learning, which will be reported in another place.

\section*{Acknowledgments}
YL is supported in part by NSF DMS-1912999 and ONR MURI N00014-20-1-2595. 

\bibliographystyle{elsarticle-num}
\bibliography{dfloss_homogenization}

\end{document}